\documentclass[letterpaper]{article} 
\usepackage{aaai25}  
\usepackage{times}  
\usepackage{helvet}  
\usepackage{courier}  
\usepackage[hyphens]{url}  
\usepackage{graphicx} 
\usepackage{natbib}  
\usepackage{caption} 
\usepackage{algorithm}
\usepackage{algorithmic}

\usepackage{newfloat}
\usepackage{listings}
\DeclareCaptionStyle{ruled}{labelfont=normalfont,labelsep=colon,strut=off} 
\floatstyle{ruled}
\newfloat{listing}{tb}{lst}{}
\floatname{listing}{Listing}
\usepackage{booktabs}
\usepackage{multirow}
\usepackage{pifont}
\newcommand{\cmark}{\ding{51}}%
\newcommand{\xmark}{\ding{55}}%
\PassOptionsToPackage{table}{xcolor}
\usepackage{amsmath}
\usepackage{amsfonts}
\usepackage{subcaption}

\title{ImagiNet: A Multi-Content Benchmark for Synthetic Image Detection}
\author{
    Delyan Boychev\textsuperscript{\rm 1}, Radostin Cholakov\textsuperscript{\rm 2}
}
\affiliations{
    \textsuperscript{\rm 1}INSAIT, Sofia University "St. Kliment Ohridski"\\ \textsuperscript{\rm 2}Stanford University \\

    delyan.boychev@insait.ai, radicho@stanford.edu
}

\usepackage{bibentry}

\begin{document}

\maketitle

\begin{abstract}
Recent generative models produce images with a level of authenticity that makes them nearly indistinguishable from real photos and artwork. Potential harmful use cases of these models, necessitate the creation of robust synthetic image detectors. However, current datasets in the field contain generated images with questionable quality or have examples from one predominant content type which leads to poor generalizability of the underlying detectors. We find that the curation of a balanced amount of high-resolution generated images across various content types is crucial for the generalizability of detectors, and introduce \textit{ImagiNet}, a dataset of 200K examples, spanning four categories: photos, paintings, faces, and miscellaneous. Synthetic images in \textit{ImagiNet} are produced with both open-source and proprietary generators, whereas real counterparts for each content type are collected from public datasets. The structure of ImagiNet allows for a two-track evaluation system: i) classification as real or synthetic and ii) identification of the generative model. To establish a strong baseline, we train a ResNet-50 model using a self-supervised contrastive objective (SelfCon) for each track which achieves evaluation AUC of up to 0.99 and balanced accuracy ranging from 86\% to 95\%, even under conditions that involve compression and resizing. The provided model is generalizable enough to achieve zero-shot state-of-the-art performance on previous synthetic detection benchmarks. We provide ablations to demonstrate the importance of content types and publish code and data.
\end{abstract}

%

\section{Introduction}

State-of-the-art generative models are rapidly improving their ability to produce nearly identical images to authentic photos and artwork. Diffusion models (DMs) \cite{ho2020denoising, rombach2022highresolution}, variational auto-encoders (VAEs) \cite{harvey2022conditional}, and generative adversarial networks (GANs) \cite{goodfellow2014generative} are being utilized in various ways to achieve data augmentation, text-to-image and image-to-image generation, inpainting and outpainting. They facilitate the production of visuals and spatial effects for downstream use in the entertainment, gaming, and marketing industries. On the other hand, these models can be misused by malicious actors \cite{masood2021deepfakes}. Thus, there is an increasing demand for improved synthetic image recognition models. Prior work \cite{wu2023generalizable, gragnaniello2021gan, corvi2022detection} employs standard classifiers but struggles with overfitting, bias, and poor generalization to novel generators, limiting effectiveness in synthetic content detection.
One key area that has yet to be fully explored in synthetic detection is the creation of training datasets with a broader range of content types and generator sources.

Previous datasets (Table~\ref{tab:prev_datasets}) primarily feature GAN-generated images and lack diversity in resolution, generator types, and content, leading to biases and overfitting issues \cite{corvi2022detection, gragnaniello2021gan, wu2023generalizable, 5995347}. \citet{rahman2023artifact} provide a diverse benchmark with multiple generators and content types, but the resized low-resolution images make it more suitable for benchmarking rather than training.

\begin{table}[t]
\centering
\begin{tabular}{@{\hskip 4pt}l@{\hskip 4pt}c@{\hskip 4pt}c@{\hskip 4pt}c@{\hskip 4pt}c@{\hskip 4pt}}
\toprule
\multicolumn{1}{c}{\textbf{Train/Eval}} & \textbf{Corvi2022} & \textbf{Wu2023} & \textbf{ArtiFact}  & \textbf{Ours} \\
\midrule
Balanced & \cmark / \xmark & \cmark / \cmark & - / \xmark & \cmark / \cmark \\ \midrule
\begin{tabular}[c]{@{}l@{}}Multiple \\ generators\end{tabular} & \xmark / \cmark & \cmark / \cmark & - / \cmark & \cmark / \cmark \\ \midrule
\begin{tabular}[c]{@{}l@{}}Proprietary \\ generators\end{tabular} & \xmark / \cmark & \xmark / \cmark & - / \xmark & \cmark / \cmark \\ \midrule
\begin{tabular}[c]{@{}l@{}}Multiple \\ content types\end{tabular} & \xmark / \xmark & \cmark / \cmark & - / \cmark & \cmark / \cmark \\ \midrule
\begin{tabular}[c]{@{}l@{}}
Synthetic\\ resolution
\end{tabular} & 
\begin{tabular}[c]{@{}c@{}}
\tiny $256 \times 256$ /\\ 
\tiny $1024 \times 1024$
\end{tabular} & 
\begin{tabular}[c]{@{}c@{}}
\tiny $1024 \times 1024$ /\\ 
\tiny $8000 \times 8000$
\end{tabular} & 
\begin{tabular}[c]{@{}c@{}}
\tiny - /\\ 
\tiny $200 \times 200$
\end{tabular} & 
\begin{tabular}[c]{@{}c@{}}
\tiny $1792 \times 1024$ /\\ 
\tiny $1792 \times 1024$
\end{tabular} \\

\bottomrule
\end{tabular}
\caption{\centering Feature comparison of previous synthetic datasets. `-' signifies that data is not available.}
\label{tab:prev_datasets}
\end{table}

We propose a new benchmark and balanced training set for synthetic image detection called \textit{ImagiNet}\footnote{ \url{https://github.com/delyan-boychev/imaginet}}. It includes images from novel open-source and proprietary generators. Our main goal is to study ways to address the challenge of generalizability by training on diverse data. The images are created by either GAN \cite{goodfellow2014generative}, DM \cite{Rombach_2022_CVPR}, or a proprietary generator -- Midjourney \cite{midjourney} or DALL·E \cite{dalle3}. Our benchmark includes two main testing tracks: synthetic image detection and model identification. Testing is performed under perturbations like JPEG compression and resizing, simulating social network conditions as in previous works \cite{corvi2022detection}. All images are high-resolution, similar to those on social networks, for more consistent results.

\section{Dataset Construction}

The \textit{ImagiNet} dataset consists of images from various open-source and proprietary image generators to encompass the distinct ``fingerprints" they impart.

\begin{table}[t]
\centering
\begin{tabular}{llll}
\toprule
\multicolumn{2}{c}{\textbf{Real}} & \multicolumn{2}{c}{\textbf{Synthetic}} \\ \midrule
\textbf{Source} & \multicolumn{1}{l|}{\textbf{Number}} & \textbf{Source} & \textbf{Number} \\ \midrule
\multicolumn{4}{c}{\textbf{Photos (30\%)}} \\ \midrule
ImageNet & \multicolumn{1}{l|}{7.5K} & StyleGAN-XL & 7.5K \\
LSUN & \multicolumn{1}{l|}{7.5K} & ProGAN* & 7.5K \\
COCO & \multicolumn{1}{l|}{15K} & SD v2.1/SDXL v1.0 & 15K \\ \midrule
\multicolumn{4}{c}{\textbf{Paintings (22.5\%)}} \\ \midrule
 & \multicolumn{1}{l|}{} & StyleGAN3 & 11.25K \\
     & \multicolumn{1}{l|}{} & SD v2.1/SDXL v1.0 & 5.625K \\
\multirow{-3}{*}{\begin{tabular}[c]{@{}l@{}}WikiArt \\ \\ Danbooru \end{tabular}} & \multicolumn{1}{l|}{\multirow{-3}{*}{\begin{tabular}[c]{@{}l@{}}11.25K\\ \\ 11.25K\end{tabular}}} & Animagine XL & 5.625K \\ \midrule
\multicolumn{4}{c}{\textbf{Faces (22.5\%)}} \\ \midrule
 & \multicolumn{1}{l|}{} & StyleGAN-XL & 11.25K \\
\multirow{-2}{*}{FFHQ} & \multicolumn{1}{l|}{\multirow{-2}{*}{22.5K}} & SD v2.1/SDXL v1.0 & 11.25K \\ \midrule
\multicolumn{4}{c}{\textbf{Uncategorized (25\%)}} \\ \midrule
 & \multicolumn{1}{l|}{} & Midjourney* & 12.5K \\
\multirow{-2}{*}{Photozilla} & \multicolumn{1}{l|}{\multirow{-2}{*}{25K}} & DALL·E 3*  & 12.5K \\ \midrule
\textbf{Total} & \multicolumn{1}{l|}{\textbf{100K}} & \textbf{Total} & \textbf{100K} \\
\bottomrule
\end{tabular}
\caption{ \centering
\textit{ImagiNet} dataset structure with two main categories and four subcategories. * signifies images sourced from public datasets.
}
\label{tab:dataset_structure}
\end{table}

\textbf{Dataset Structure} (Table~\ref{tab:dataset_structure}) -- The dataset structure is designed to represent real-world scenarios where images of different content types might be used. \textit{ImagiNet} examples are split into two main categories -- real and synthetic images. To mitigate content-related biases, the dataset is divided into four subcategories -- photos, paintings, faces, and miscellaneous. Such images are commonly found on the World Wide Web and are the main subject of generative applications. We provide a balanced amount of synthetically generated images and real counterparts in each subcategory. The source datasets and generator models are given in Table~\ref{tab:dataset_structure}. Images from models marked with * are sourced as follows: ProGAN from \citet{wang2019cnngenerated}, Midjourney from \citet{pan2023journeydb}, DALL·E 3 from LAION \cite{laiondalle3}; in addition we generated 800 DALL·E 3 images to reach our desired dataset size. Synthetic groups are generated with pre-trained models: GAN images are labeled as GAN, Stable Diffusion as SD, and proprietary models as standalone.

\textbf{Real Images Sampling} -- The real images are randomly sampled from each real counterpart dataset described in Table~\ref{tab:dataset_structure} \cite{russakovsky2015imagenet, yu2016lsun, lin2015microsoft, artgan2018, danbooru2021, karras2019stylebased, singhal2021photozilla}. The images in our test set are sampled from the validation and testing splits of these sets.

\begin{figure}[t]
    \centering
    \begin{subfigure}{0.95\linewidth}
    \centering
    \includegraphics[width=1\linewidth]{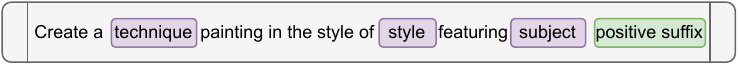}
    \caption{Painting Generation}
    \label{fig:painting_prompt}
    \end{subfigure}
    \\
    \begin{subfigure}{\linewidth}
    \centering
    \includegraphics[width=0.95\linewidth]{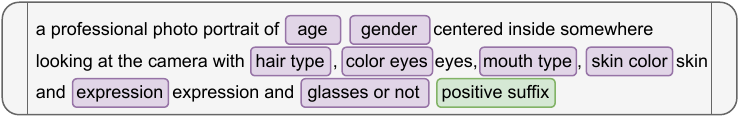}
    \caption{Face Generation}
    \label{fig:face_prompt}
    \end{subfigure}
    \caption{\centering Prompt structures for image generation.}
\end{figure}

\textbf{Image Generation Procedure} -- To generate images with GANs (StyleGAN-XL \cite{sauer2022styleganxl}, StyleGAN3 \cite{Karras2021}), we sample random latent code (it is selected according to model requirements) for a given seed and feed the generator with it unconditionally. For DMs and private generators (SD v2.1 \cite{Rombach_2022_CVPR}, SDXL v1.0 \cite{podell2023sdxl}, Animagine XL \cite{animaginexl}, DALL·E 3 \cite{dalle3}), however, textual guidance is needed, thus we first search manually for appropriate negative prompts and positive suffixes to increase the quality of the produced images. The construction of each prompt is in descriptive form. For photos, we utilize the captions from COCO \cite{chen2015microsoft} to prompt the generators and achieve images with sufficient quality. For paintings, instead of using a pre-generated set of captions for prompting, we create lists of styles, techniques, and subjects with GPT-3.5 Turbo \cite{brown2020language}. After that, we fit these characteristics of the paintings in a descriptive sentence shown in Figure~\ref{fig:painting_prompt}, which guides the model to generate varied images. The gaps are filled respectively with an item from the given list, and in the end, a positive suffix is added. The procedure for face generation is similar -- Figure~\ref{fig:face_prompt} presents the structure of the prompt. All the lists for filling in the guiding instructions, as well as the positive suffixes and negative prompts. The last model AnimagineXL, a fine-tuned SDXL \cite{podell2023sdxl} variant for art generation, uses only tags from the Danbooru dataset \cite{danbooru2021} for prompting.

\textbf{Dataset Splits} -- From the whole set, we sample 80\% of the images from each category and subcategory with an equal number of images from the different generators. The number of images in the training set is 160K, respectively 40K are left for testing. We aim to provide a balanced (an equal number of images for each model) calibration set sampled from the training set. It consists of 80K examples in total.

\textbf{Labelling and Evaluation Tracks} -- All the images of the dataset are labelled. They have four labels -- source (real or synthetic), content type, generator group (e.g., GAN), and specific generator (e.g., ProGAN). In our benchmark, we have two tracks -- synthetic image detection and model identification. Perturbations are applied on the test set to simulate social network conditions \cite{corvi2022detection}. First, we do a large square crop (ranging from 256 to the smaller dimension of the image) of the image and, after that, resize it to $256 \times 256$. After that, we compress 75\% of the images with JPEG or WebP compression.

\textbf{Dataset Access} -- We provide the synthetic images we generated for this work, along with those from DALL·E 3, which are collected under a Creative Commons Zero license. Both the real counterparts and the additional part of synthetic content (Midjourney and ProGAN examples) can be downloaded from their sources. The whole dataset can be reconstructed with the scripts in our repository, which also includes the list of sources and our synthetic data.

\section{Baseline Training}
To train our baseline, we initialize a ResNet-50 model with pre-trained ImageNet weights and modify its early layers to avoid downsampling, following \citet{gragnaniello2021gan}.

In the first stage of training, we train a backbone with a contrastive objective $\mathcal{L}_{SC}$, as proposed by \citet{bae2022selfcontrastive}:

\begin{align}
    \mathcal{L}_{SC} &= \sum\limits_{\substack{i \in A \\ \omega \in \Omega}} 
    \frac{-1}{\lvert P(i) \rvert \lvert \Omega \rvert} \nonumber \\
    &\quad\sum\limits_{\substack{p \in P(i) \\ \omega' \in \Omega}} 
    \log \frac{\mathrm{exp}(\omega(x_{i}) \cdot \omega'(x_{p}) / \tau)}
    {\sum\limits_{l \in Q(i)} \mathrm{exp}(\omega(x_{i}) \cdot \omega'(x_{l}) / \tau)} 
\end{align}

\noindent where $A \equiv \{1, ..., N\}$ is a set of indices for all batch examples, $Q(i) \equiv A \backslash \{i\}$ (similarity between $z_{i}$ and $z_{i}$ is redundant), and $P(i) \equiv \{p \in Q(i): \hat{y}_p = \hat{y}_i\}$ is the set of positive examples for a given example $i$.

A sub-network is attached to the backbone. Its main responsibility is to produce an alternative view of the images in the latent space instead of additional augmented samples to design the SelfCon loss with a single-viewed (augmented once) batch. The sub-network could be a fully connected layer or another architecture with the same function as the backbone. The sub-net $H_{sub}(.)$ is attached to the backbone and projects the latent representations $F_m(.)$ obtained after the $m$-th ResNet block. The network has two output mapping functions  $\Omega \equiv \{ H_{sub}(F_{m}(.)), H(F(.))\}$ for a given input $x_{i}$. In our case, the mapping functions $H(.)$ and $H_{sub}(.)$ output representations in $\mathbb{R}^{128}$. This involves accumulating $\mathcal{L}_{SC}$ applied on two labellings - synthetic detection and model identification labels, with each loss assigned equal weight. When optimizing the model detection objective, real images in the batch are ignored.  
To address the increased memory demands of removing downsampling in early ResNet-50 layers and the large batch size requirements of SelfCon, we adopt gradient caching \cite{gao2021scaling}, a technique initially designed for language model contrastive losses. We modify it for use with SelfCon \cite{bae2022selfcontrastive}, SupCon \cite{khosla2021supervised}, and SimCLR \cite{chen2020simple}. This approach calculates the loss on the entire batch but accumulates gradients in smaller chunks, allowing for large batch sizes and efficient training on memory-constrained GPUs.

The second stage involves calibrating the model. We detach the sub-network and projection heads, replacing the output projection head with a multilayer perceptron classifier. This classifier is then trained using cross-entropy loss on a balanced dataset to perform both origin and model detection. We update the batch normalization statistics within the backbone's residual blocks, following \citet{schneider2020improving}, to enhance robustness against perturbations not encountered during pre-training.

\section{Experiments and Results}

\begin{figure}[t]
    \centering
    \begin{subfigure}{0.75\linewidth}
    \includegraphics[width=\linewidth]{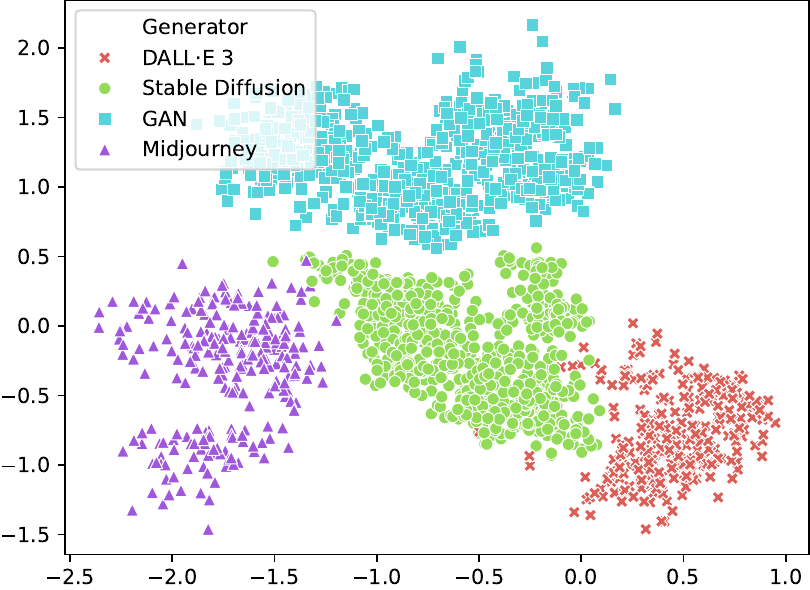}
    \end{subfigure}
    \caption{\centering Dimensionality reduction vizualization of the backbone representations for a subset of \textit{ImagiNet}.}
    \label{fig:dim_red_plots}
\end{figure}

\begin{table*}[t]
\label{tab:imaginet_tab}
\centering
\begin{tabular}{l|c|c|c|c|c}
\toprule
\multicolumn{1}{c|}{\textbf{ACC / AUC}} & \textbf{Grag2021} & \textbf{Corvi2022} & \textbf{Wu2023} & \textbf{Corvi2022}* & \textbf{Ours}* \\ \midrule
GAN                                     & 0.6889 / 0.8403   & 0.6822 / 0.8033    & 0.6508 / 0.6971 & 0.8534 / 0.9416                                    & \textbf{0.9372} / \underline{0.9886}                                  \\
SD                                      & 0.5140 / 0.5217   & 0.6112 / 0.6851    & 0.6367 / 0.6718 & 0.8693 / 0.9582                                    & \textbf{0.9608} / \underline{0.9922}                                 \\
Midjourney                              & 0.4958 / 0.5022   & 0.5826 / 0.6092    & 0.5326 / 0.5289 & 0.8880 / 0.9658                                    & \textbf{0.9652} / \underline{0.9949}                                  \\
DALL·E 3                                & 0.4128 / 0.3905   & 0.5180 / 0.5270    & 0.5368 / 0.5482 & 0.8906 / 0.9759                                  & \textbf{0.9724} / \underline{0.9963}                                  \\ \midrule
\textbf{Mean}                           & 0.5279 / 0.5637   & 0.5985 / 0.6562    & 0.5892 / 0.6115 & 0.8753 / 0.9604                                    & \textbf{0.9589} / \underline{0.9930}     \\ \bottomrule                     
\end{tabular}
\caption{\centering
\textit{ImagiNet} test set evaluation -- best \textbf{ACC}/\underline{AUC}. * means trained on \textit{ImagiNet}.
}
\label{tab:imaginet_res}
\end{table*}

\begin{table}[t]
\centering

\begin{tabular}{l|c|c}
\toprule
\multicolumn{1}{c|}{\textbf{ACC / AUC}} & \textbf{Wu2023} & \textbf{Ours}* \\ \midrule
DreamBooth                              & 0.9049 / 0.9733 & \textbf{0.9601} / \underline{0.9950} \\
MidjoruneyV4                            & 0.8907 / 0.9495 & \textbf{0.9675} / \underline{0.9959} \\
MidjourneyV5                            & 0.8540 / 0.9224 & \textbf{0.9745} / \underline{0.9991} \\
NightCafe                               & \textbf{0.8962} / \underline{0.9652} & 0.8931 / 0.9644 \\
StableAI                                & 0.8806 / 0.9534 & \textbf{0.9574} / \underline{0.9947} \\
YiJian                                  & 0.8392 / 0.9233 & \textbf{0.9045} / \underline{0.9726} \\ \midrule
\textbf{Mean}                           & 0.8776 / 0.9479 & \textbf{0.9428} / \underline{0.9870} \\ \bottomrule
\end{tabular}
\caption{\centering
Practical test set \cite{wu2023generalizable} evaluation -- best \textbf{ACC}/\underline{AUC}. * means trained on \textit{ImagiNet}.
}
\label{tab:practical_tab}
\end{table}

\begin{table}[t]
\centering \fontsize{7pt}{8pt}\selectfont
\begin{tabular}{l|c|c|c}
\toprule
\multicolumn{1}{c|}{\textbf{ACC / AUC}} & \textbf{Corvi2022} & \textbf{Corvi2022}* & \textbf{Ours}* \\ \midrule
ProGAN                                  & 0.9117 / 0.9994    & 0.9030 / 0.9995                                    & 0.8974 / 0.9991                                  \\
StyleGAN2                               & 0.8662 / 0.9455    & 0.8675 / 0.9479                                    & \textbf{0.8884} / 0.9759                                     \\
StyleGAN3                               & 0.8557 / 0.9416    & 0.8705 / 0.9440                                    & \textbf{0.8824} / \underline{0.9707}                                  \\
BigGAN                                  & \textbf{0.8952} / 0.9699    & 0.8980 / 0.9882                                    & 0.8934 / 0.9864                                  \\
EG3D                                    & \textbf{0.9062} / 0.9756    & 0.8450 / 0.9160                                    & 0.8964 / \underline{0.9913}                                  \\
Taming Tran                             & \textbf{0.9112} / \underline{0.9902}    & 0.8538 / 0.9278                                    & 0.8829 / 0.9651                                  \\
DALL·E Mini                             & \textbf{0.9117} / \underline{0.9914}    & 0.9015 / 0.9792                                    & 0.8924 / 0.9786                                  \\
DALL·E 2                                & 0.6507 / 0.7590    & 0.7370 / 0.8302                                    & \textbf{0.7729} / \underline{0.8590}                                  \\
GLIDE                                   & \textbf{0.9062} / \underline{0.9780}    & 0.8730 / 0.9429                                    & 0.8539 / 0.9347                                  \\
Latent Diff                             & \textbf{0.9117} / \underline{0.9998}    & 0.9017 / 0.9989                                    & 0.8959 / 0.9902                                  \\
Stable Diff                             & \textbf{0.9117} / \underline{0.9999}    & 0.9030 / 0.9998                                    & 0.8969 / 0.9956                                  \\
ADM                                     & \textbf{0.7927} / \underline{0.8772}    & 0.7875 / 0.8710                                    & 0.7704 / 0.8550                                  \\ \midrule
\textbf{Mean}                           & \textbf{0.8692} / 0.9523    & 0.8618 / 0.9446                                    & 0.8686 / \underline{0.9585}                                  \\ \bottomrule
\end{tabular}

\caption{\centering
Corvi test set \cite{corvi2022detection} evaluation -- best \textbf{ACC}/\underline{AUC}. * means trained on \textit{ImagiNet}.
}
\label{tab:corvi_tab}
\end{table}

First, we evaluate the described baseline against existing synthetic datasets. Then, we examine the importance of balancing content types in \textit{ImagiNet} for the performance of detectors.

\textbf{Baseline} -- During the first stage, the backbone is optimized with SGD \cite{ruder2017overview} for 400 epochs with batch size $N=200$ on the \textit{ImagiNet} training set. The initial learning rate of $0.005$ is warmed up linearly \cite{ma2021adequacy} for 10 epochs and is cosine annealed \cite{loshchilov2017sgdr} afterwards. The second stage continues for 5 epochs with AdamW optimizer \cite{loshchilov2017sgdr} and constant learning rate $0.0001$, weight decay $0.001$, $\beta_{1} = 0.9$ and $\beta_{2} = 0.99$. After the pre-training procedure, we visualize the model representations of the images in the test set by applying Autoencoder dimensionality reduction \cite{Meng_2017}. The plots in Figure~\ref{fig:dim_red_plots} show the ability of our model to cluster each generator's images.

As shown in Table~\ref{tab:imaginet_res}, the baseline achieves AUC of up to 0.99 and balanced accuracy over 95\% on \textit{ImagiNet}. To demonstrate its generalization abilities we evaluate zero-shot performance on the datasets from \cite{wu2023generalizable} in Table~\ref{tab:practical_tab}, and \cite{corvi2022detection} in Table ~\ref{tab:corvi_tab}. Our baseline is able to outperform the original method of Wu2023 and remains comparable on Corvi2022's benchmark. The baseline shows a substantial improvement of 12\% in ACC on DALL·E 2 examples since it is trained on DALL·E 3 images. The results on StyleGAN3 and StyleGAN2 are increased by 1-2\%. Table~\ref{tab:inference_time} presents a comparison of the inference time of our detector with previous models. We also train the model proposed in Corvi2022 on \textit{ImagiNet} to demonstrate that the balanced dataset elicits generalizable performance regardless of the architecture and training procedure.

\textbf{Content Type Balancing} -- To investigate the influence of specific content types and identify potential biases, we conducted an ablation study inspired by Leave-One-Out Cross-Validation (LOOCV). Separate models were trained, each with one content type excluded from its training data, while maintaining equal training data overall. The isolation of the specific category influence allows us to identify potential biases through drastic changes in performance when tested on the unseen group of examples.

From the synthetic images in our \textit{ImagiNet} dataset, we focused on those generated by Stable Diffusion due to its presence in all image subcategories, thus eliminating potential generator-specific biases. We sampled a balanced subset containing 4500 real and 4500 synthetic (Stable Diffusion only) images per subcategory (photos, paintings, faces). For each model, we used a ResNet-18 architecture, training it from scratch for 200 epochs to avoid any biases from pre-trained models. Each model was trained on 18000 images with one category left out. For evaluation, we sample 1000 real and 1000 synthetic images for each category.\footnote{Our analysis revealed no significant bias toward the resolution of real images across different content type groups.}

\begin{figure}[t]
    \centering
    \begin{subfigure}{\linewidth}
        \includegraphics[width=\linewidth]{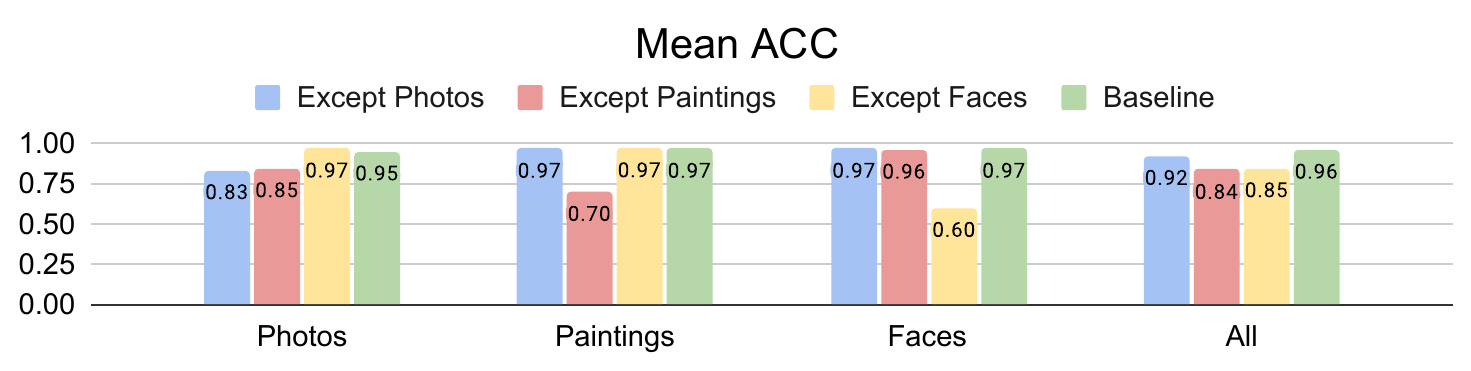}
    \end{subfigure}
    \begin{subfigure}{\linewidth}
        \includegraphics[width=\linewidth]{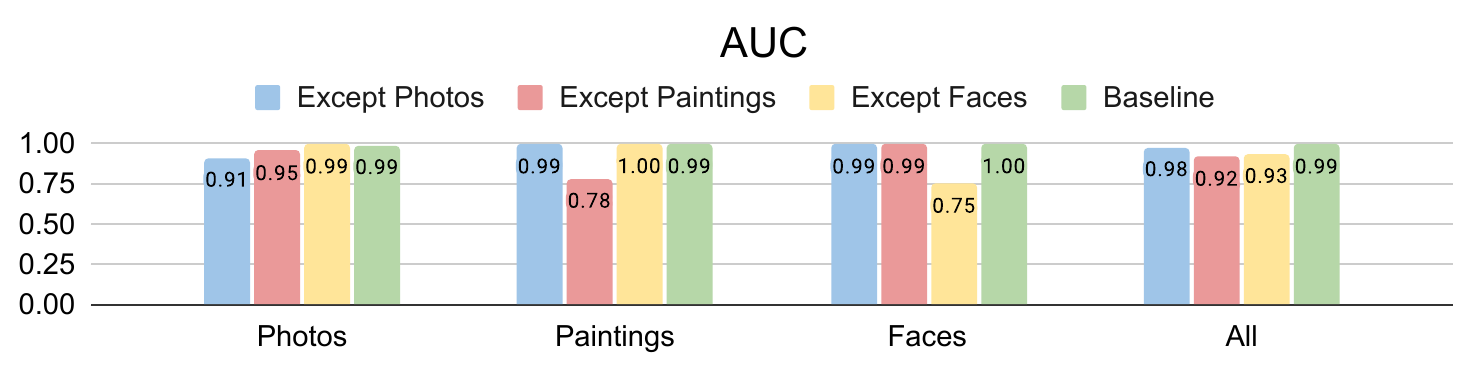}
    \end{subfigure}
    \caption{\centering Mean accuracy and AUC on the different models trained by leaving one content type out.}
    \label{fig:mean_acc_auc_ct}
\end{figure}

Figure~\ref{fig:mean_acc_auc_ct} demonstrate that models trained by excluding a specific content type exhibit overfitting and generally lower synthetic accuracy when tested on that content type. Notably, the ``Except Faces" model overfits the real image distribution, suggesting that bias is introduced not only by synthetic images but also by real images. The AUC plot in Figure~\ref{fig:mean_acc_auc_ct} reveals high variance from expected values for the ``Except Painting" and ``Except Faces" models on their respective content types, highlighting the inability to distinguish between the real and synthetic classes at all possible thresholds. This suggests that training on diverse content types is essential for mitigating bias. The baseline model, trained on all types, does not overfit on the test set.


\begin{table}[t]
\centering
\begin{tabular}{l|l|l|l}
\toprule
\textbf{Grag2021} & \textbf{Corvi2022} & \textbf{Wu2023} & \textbf{Ours} \\ \midrule
24.30             & 49.53              & \textbf{16.01}  & 25.10         \\ \bottomrule
\end{tabular}

\caption{
\centering
Inference time in milliseconds for $448\times448$ image on RTX 4090 GPU.
}
\label{tab:inference_time}

\end{table}

\section{Conclusion}

In this work: (1) we demonstrate the importance of balancing content types in synthetic image datasets; (2) we provide a modest-in-size but high quality benchmark for training and evaluating synthetic detectors; (3) we provide a strong baseline which generalizes on third-party datasets.
\section{Acknowledgements}
We would like to thank Sangmin Bae for the advice provided. We also acknowledge the America for Bulgaria and Beautiful Science foundations for partially funding the computational resources used as a part of the Diffground OSS project. The Google ML Developer Programs team supported this work by providing Google Cloud and Colab credits. This research was partially supported by the Bulgarian National Program "Education with Science" and by the Ministry of Education and Science of Bulgaria (support for INSAIT, part of the Bulgarian National Roadmap for Research Infrastructure).

\bibliography{aaai25}
\newpage
\appendix
\section{Model Identification Track}

\textbf{Testing Settings} -- For this evaluation track, we use only the synthetic images with the same applied perturbations from the synthetic track. 

\textbf{Discussion} -- Interestingly, even aggressive perturbations such as resize and compression do not significantly harm the performance of the model identifier (Table~\ref{tab:model_identification}). The results confirm that each generator has its unique "fingerprints", which are studied in other works \cite{corvi2022detection}. The model identification track is open for evaluation of other novel detectors.
\subsection{Model Identification Under Perturbations}
 We investigate the robustness of our model's classification performance under image perturbations, specifically JPEG compression and image resizing with linear scaling of the crop size.

\begin{figure}[!t]
     \centering
     \caption{\centering Accuracy of model identification classifier under perturbations.}
     \begin{subfigure}{0.9\linewidth}
     \centering
     \includegraphics[width=\linewidth]{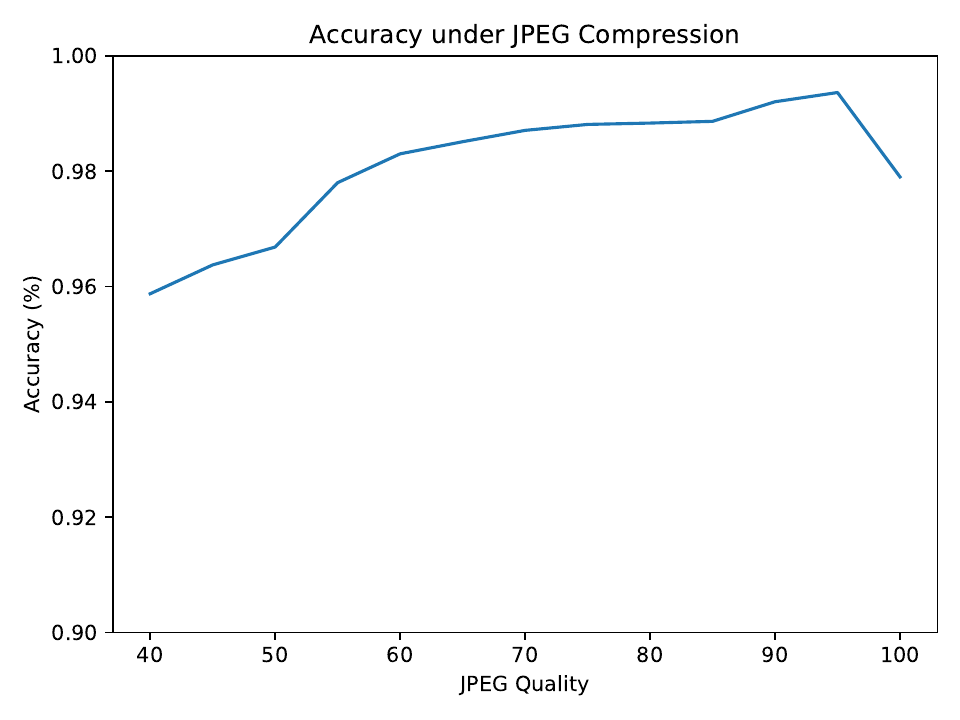}
     \caption{JPEG compression}
     \label{fig:jpeg}
     \end{subfigure}
     \begin{subfigure}{0.9\linewidth}
     \centering
     \includegraphics[width=\linewidth]{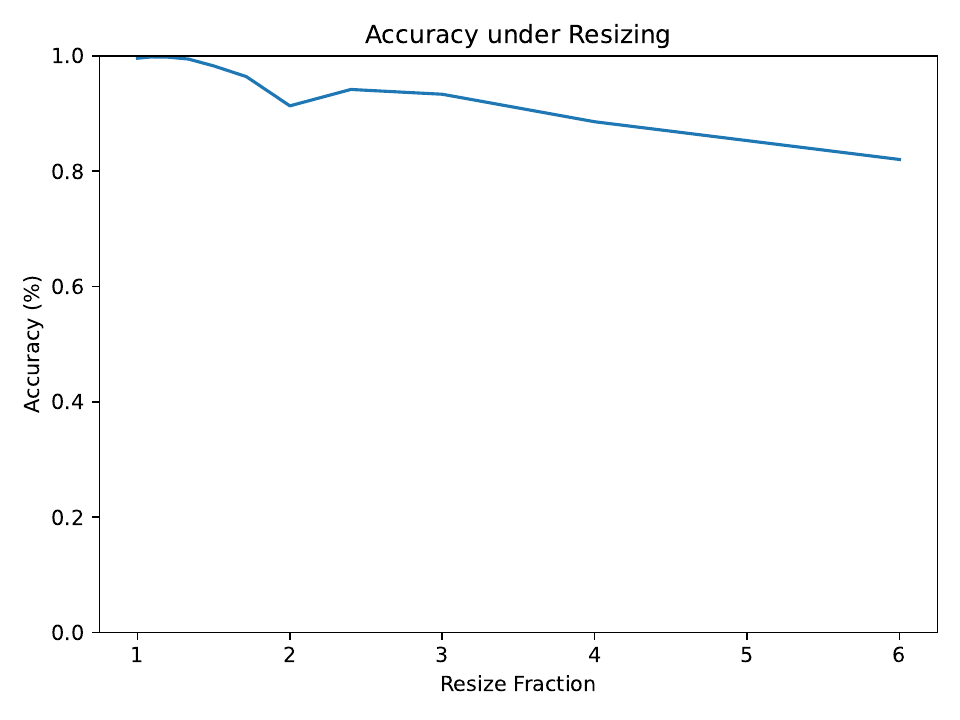}
     \caption{Resizing}
     \label{fig:resize}
     \end{subfigure}
\end{figure}

 \textbf{JPEG Compression} -- The impact of JPEG compression on model performance is presented in Figure~\ref{fig:jpeg}. The results indicate that even aggressive JPEG compression with a quality factor as low as 40 does not significantly degrade model performance.  This suggests a level of resilience to the artefacts introduced by JPEG compression within our classification model.

 \textbf{Image Resizing} -- Figure~\ref{fig:resize} illustrates the model's performance under image resizing with linear scaling of the crop size. To prepare the images for analysis, we first apply a center crop with dimensions $256r \times 256r$, where $r$ denotes the scaling fraction along the x-axis of the plot. Subsequently, the images are resized to a standard resolution of $256 \times 256$ pixels. Similarly, to JPEG compression we have a slight degradation in performance, but with much aggressive perturbation. This suggests that the model is robust to one of the most common augmentations applied to images in social networks.




\begin{table}[!t]
\centering
\caption{\centering Results of our detector on ImagiNet model identification track - ACC. The AUC is $0.9980$}
\label{tab:model_identification}
\begin{tabular}{cccc|c}
\hline
\textbf{GAN} & \textbf{SD} & \textbf{Midjourney} & \textbf{DALL·E 3} & \textbf{Mean} \\ \hline
0.9968 & 0.9683 & 0.9228 & 0.9160 & 0.9510 \\ \hline
\end{tabular}
\end{table}
\section{Visualizations via Relational Dimensionality Reduction}
High-dimensional data often poses challenges for analysis since relationships between data points can't be visualized and interpreted directly. To address this, we employ dimensionality reduction techniques. Specifically, we use an autoencoder architecture \cite{bank2021autoencoders} to project the data into a lower-dimensional space, allowing us to potentially identify decision boundaries. The model consists of two parts: an Encoder $\mathrm{g}_{enc}()$ and a Decoder $\mathrm{g}_{dec}()$. Each layer in the network is linear with ReLU activation. Batched vectors $R_{in} \in \mathbb{R}^{b \times n}$ - where $b$ is the batch size - are inputted to the model. The dimensionality $n$ decreases layer-wise until it reaches dimension $h$ - dimensionality of encoded vectors $R_{enc} \in \mathbb{R}^{h}$. In our case, $h=2$ since we want to plot the vectors as points in 2D space. Following that, the decoder increases the dimensionality, outputting reconstructed vector $R_{out} \in \mathbb{R}^{n}$. To achieve accurately encoded vectors, we need to minimize the Mean Squared Error (MSE) between $R_{in}$ and $R_{out}$. Due to the inability of reconstruction error to preserve the relations between vectors in the high-dimensional space, we introduce relational loss. This objective preserves the pairwise distances between vectors in the batch after decoding and also in lower-dimensional mapping. Our method adopts a similar framework to \citet{Meng_2017}. Instead of calculating the MSE between Gram matrices, we focus directly on the pairwise $l_{2}$ distances between original input vectors and their decoded counterparts, as this preserves the magnitude of the distances and allows us to maintain the original positional relationships in the data. We then calculate the MSE over these pairwise distances, the value of which indicates the average squared error in reconstructing the individual relationships present within the original data. Hence, the objective is the following:
\begin{equation}
   \frac{\sum_{i=1}^b \sum_{j=1}^b ({\lVert R_{out[i, :]} - R_{out[j, :]} \rVert}^2 - {\lVert R_{in[i, :]} - R_{in[j, :]} \rVert}^2)^2}{b^2},
\end{equation}
where $b^2$ is the number of pairs of vectors within the batch. Then, the whole objective is as follows:

\begin{equation}
\begin{split}
\mathcal{L}_{rdra} = \sum_{i=1}^{b}\sum_{j=1}^{b}\frac{(1-\alpha)}{b^2}(R_{in[i, j]} - R_{out[i, j]})^2 + \\ \frac{\alpha}{b^2}({\lVert R_{out[i, :]} - R_{out[j, :]} \rVert}^2 -  {\lVert R_{in[i, :]} - R_{in[j, :]} \rVert}^2)^2
\end{split}
\end{equation} 
The losses are scaled by hyperparameters $(1-\alpha)$ and $\alpha$, where alpha is between 0 and 1. Alpha controls the trade-off between reconstruction accuracy and preservation of pairwise distances. To find the best value, we apply hyperparameter optimization for each specific set by finding the lowest absolute error and lowest cosine distance for a set of $\alpha \in [0, 1]$.

\section{Image Generation and Prompt Engineering}
\begin{figure}[!t]
    \centering
    \includegraphics[width=0.9\linewidth]{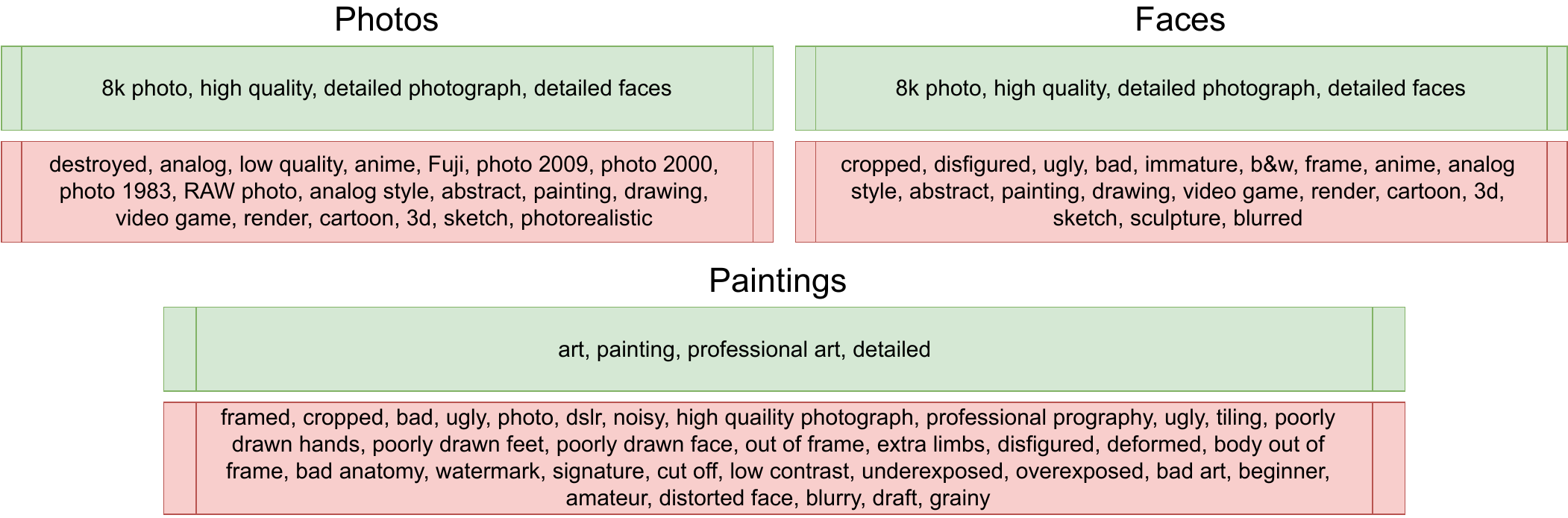}
    \caption{Positive suffixes (green) and negative prompts (red) utilized for the generation of all generative models requiring prompts.}
    \label{fig:suffix_neg_prompt}
\end{figure}

All the positive suffixes and negative prompts (presented in Figure~\ref{fig:suffix_neg_prompt}) are optimized manually by analyzing the quality of the images and how well the generative model follows the instructions. We also provide the list of guiding words for the generative model, which are filled into the original prompts provided in the main work. For painting generation, we use these values:
\begin{itemize}
    \item \textbf{technique} -- oil, watercolor, acrylic, digital art, pen and ink;
    \item \textbf{style} -- impressionism, abstract, realism, cubism, surrealism, pop art, expressionism, minimalism, post-impressionism, art deco, fauvism, romanticism, baroque, neoclassicism, surreal abstraction, hyperrealism, symbolism, pointillism, suprematism, constructivism, japan art, ukiyo-e, kinetic art, street art, digital art, naïve art, primitivism, abstract expressionism, conceptual art, futurism, precisionism, social realism, magical realism, cubofuturism, lyrical abstraction, tenebrism, synthetic cubism, metaphysical art, graffiti art, videogame art;
    \item \textbf{subject} -- landscape, portrait, still life, cityscape, abstract composition, wildlife, fantasy, architecture, seascape, flowers, people, animals, food, music, dance, sports, mythology, history, technology, science, nature, celebrity, space, transportation, underwater, emotion, dreams, folklore, literature;
\end{itemize}

For face generation, we utilize the following values:
\begin{itemize}
    \item \textbf{age} -- baby\footnote{When generating faces of babies, we exclude all the facial characteristics since they are not developed. We prompt the model only with the gender and skin color.}, young, middle-aged, elderly
    \item \textbf{gender} -- male, female
    \item \textbf{hair type} -- straight, wavy, curly
    \item \textbf{eyes} -- small, large, almond-shaped, round
    \item \textbf{mouth} -- thin lips, full lips, wide mouth, narrow mouth
    \item \textbf{expression} -- neutral expression, smiling, serious, surprised, angry
    \item \textbf{skin color} -- fair, olive, pale, medium, dark
    \item \textbf{glasses} - with glasses, without glasses
\end{itemize}

All the images are generated with different random seeds for the initial noise to achieve diverse generated content. To remove potential bias, during the image generation, we utilize different schedulers for generation \cite{chen2023importance} -- Euler Discrete, Euler Ancestral \cite{karras2022elucidating}, DPM-Solver \cite{lu2022dpmsolver}, PNDM \cite{liu2022pseudo}. 

\section{Additional Training Details}
During the training procedure, the following augmentations are utilized\footnote{No resizing of images is conducted during training.}:
\begin{enumerate}
    \item Pad if needed to $96 \times 96$ with reflection borders 
    \item Random crop $96 \times 96$
    \item 50\% of the images are perturbed by randomly selecting one of the following:
    \begin{itemize}
        \item JPEG compression with quality $Q \in [50, 95]$
        \item WebP compression with quality $Q \in [50, 95]$
        \item Gaussian blur with kernel size $k \in [3, 7]$ and standard deviation $\sigma = 0.3*((k-1)*0.5 - 1) + 0.8$
        \item Gaussian noise with variance $\sigma^{2} \in [3, 10]$
    \end{itemize}
    \item 33\% of the images are rotated to $90^{\circ}$
    \item 33\% of the images are flipped (either horizontally or vertically)
\end{enumerate}
The calibration set consists of an equal number of images from each generator and their real counterparts. During the calibration, we apply the following augumentations:
\begin{enumerate}
    \item 50\% of the images are perturbed:
    \begin{enumerate}
        \item 70\% are perturbed by randomly selecting one of the following:
        \begin{itemize}
        \item JPEG compression with quality $Q \in [50, 95]$
        \item WebP compression with quality $Q \in [50, 95]$
        \item Gaussian blur with kernel size $k \in [3, 7]$ and standard deviation $\sigma = 0.3*((k-1)*0.5 - 1) + 0.8$
        \item Gaussian noise with variance $\sigma^{2} \in [3, 10]$
    \end{itemize}
    \item 30\% are augmented with all applied in the same order:
        \begin{enumerate}
            \item Pad if needed to $256 \times 256$ with reflection borders 
            \item Random resized crop $256 \times 256$, scale $S \in [0.08, 1]$ and ratio $R \in [0.75, 1.33]$
            \item 50\% are augmented by randomly selecting one of the following:
            \begin{itemize}
                \item JPEG compression with quality $Q \in [50, 95]$
                \item WebP compression with quality $Q \in [50, 95]$
            \end{itemize}
        \end{enumerate}
    \end{enumerate}
    \item Pad if needed to $256 \times 256$ with reflection borders 
    \item Random crop $256 \times 256$
    \item 33\% of the images are rotated to $90^{\circ}$
    \item 33\% of the images are flipped (either horizontally or vertically)
\end{enumerate}

\end{document}